\documentclass[preprint,12pt,authoryear]{elsarticle}
\usepackage[ruled]{algorithm2e}
\usepackage{amsmath}
\usepackage{varwidth}
\usepackage{amssymb}
\usepackage{mathtools, cuted}
\usepackage{amsthm}

\newtheorem{theorem1}{Special Theorem}
\newtheorem{definition}[theorem1]{Definition}

\usepackage{caption}
\usepackage{subcaption}
\usepackage{booktabs}
\usepackage{graphicx}
\usepackage{float}

\usepackage{tikz}
 \usetikzlibrary{arrows,shapes,positioning,shadows,trees}
\tikzset{
  basic/.style  = {draw, text width=2cm, drop shadow, font=\sffamily, rectangle},
  root/.style   = {basic, thin, align=center, 
                   fill=white!30, text width=5em},
  level 2/.style = {basic,  thin,align=center, fill=white!60,
                   text width=5em},
  level 3/.style = {basic, thin, align=left, fill=white!60, text width=5em}
}

\begin{document}

\begin{frontmatter}

\title{Aggregating Predictions on Multiple Non-disclosed Datasets using Conformal Prediction}

\author[label1]{Ola Spjuth}
\ead{ola.spjuth@farmbio.uu.se}
\author[label2,label3]{Lars Carlsson}
\ead{Lars Carlsson <dr.lars.carlsson@gmail.com}
\author[label1]{Niharika Gauraha}
\ead{niharika.gauraha@farmbio.uu.se}

\address[label1]{Department of Pharmaceutical Biosciences \\
       Uppsala University\\
       Uppsala, Sweden}
\address[label2]{Department of Computer Science, \\Royal Holloway, University of London, \\Egham Hill, Egham, Surrey, United Kingdom}
\address[label3]{Stena Line, Gothenburg, Sweden}

\begin{abstract}
Conformal Prediction is a machine learning methodology that produces valid prediction regions under mild conditions. 
In this paper, we explore the application of making predictions over multiple data sources of different sizes without disclosing data between the sources.
We propose that each data source applies a transductive conformal predictor independently using the local data, and that the individual predictions are then aggregated to form a combined prediction region. We demonstrate the method on several data sets, and show that the proposed method produces conservatively valid predictions and reduces the variance in the aggregated predictions. We also study the effect that the number of data sources and size of each source has on aggregated predictions, as compared with equally sized sources and pooled data.
\end{abstract}

\end{frontmatter}

\section{Introduction}
%it is not uncommon to have multiple sources of data .

\begin{figure}[b!]
    \includegraphics[width=0.95\textwidth]{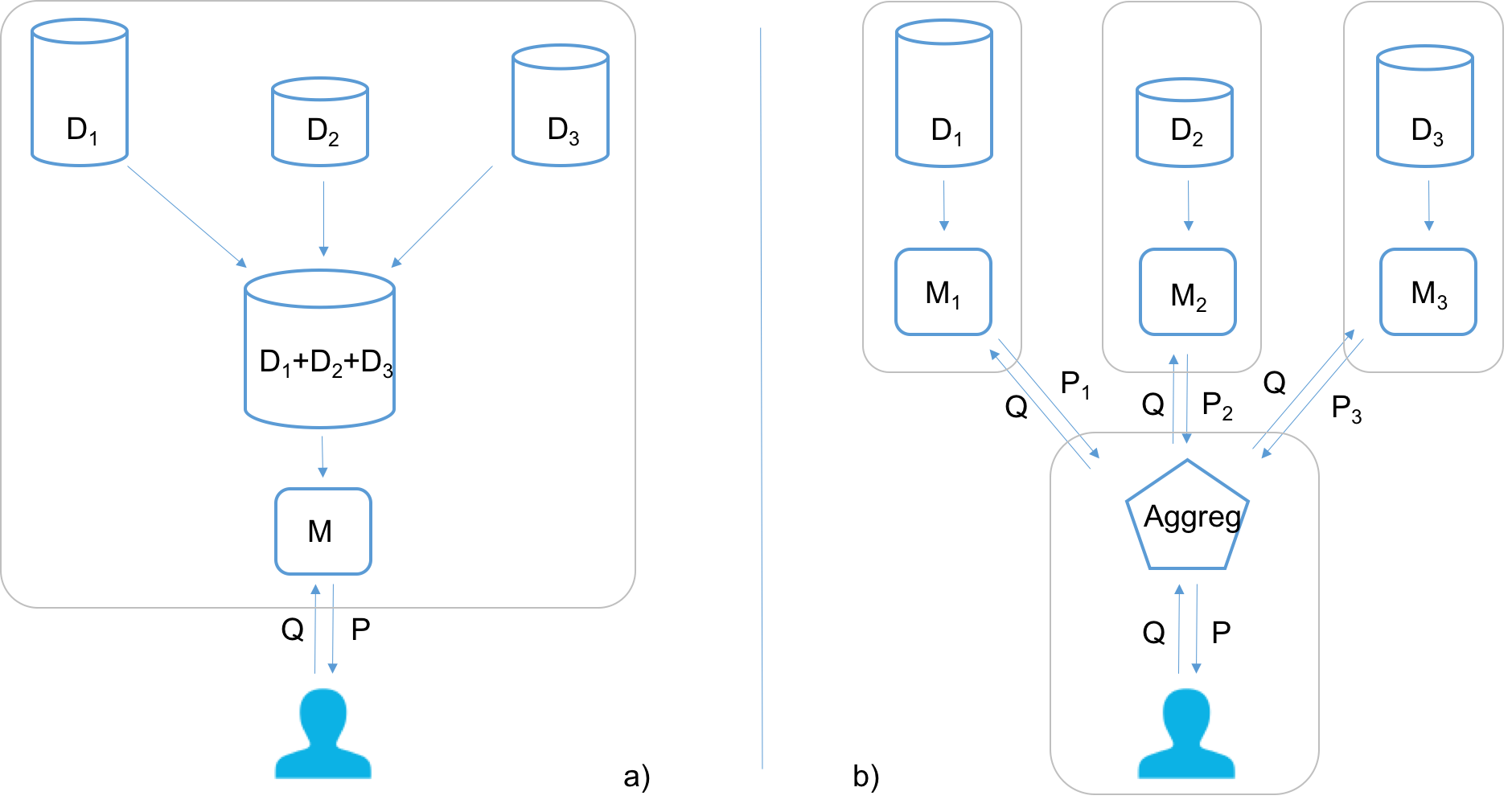}
    \caption{\textbf{a}) The most common approach is to collect data from different data sources ($D_1-D_3$) into a single dataset, which then is used to train a model $M$ that can be used to make a prediction $P$ on a query object $Q$. \textbf{b)} The aggregated TCP approach implies that a model $M_n$ is trained at each data source $D_n$, and the query object Q is passed on to each model, and predictions $P_n$ are then aggregated to deliver the resulting prediction $P$. The gray wireframes are used to visualize the different actors taking part in the procedure independent of each other.}
  \label{fig:overview}
\end{figure}

The increasing volumes of data generated in virtually all scientific and industrial domains presents formidable challenges to store and analyze. Of particular interest is to make use of the information in statistical learning systems with the objective to make predictions on future objects. If data reside in multiple data sources, possibly in different databases or geographical locations, the most common approach is to collect all data intended for model building in a single location, such as a data warehouse or a file system, after which it is subjected to learning algorithms and subsequent predictions (see Figure~\ref{fig:overview}a). However, if data is large or if the data owners do not allow such pooling of data, this strategy may not be possible. One example is in the pharmaceutical industry where large databases are available at companies, each holding results on e.g. chemical compounds tested against different endpoints in drug discovery projects. This information is valuable and sensitive for these companies, but at the same time there are occasions where companies would want to contribute to predictive models without disclosing the data to others, such as in collaborative efforts or precompetitive alliances. There are approaches that have been developed towards integrated analysis that do not require sharing of original data, but these come with limitations. For example, methods for integrated analysis of non-sensitive availability data derived from original data has been developed~\citep{Spjuth:2016ly} but these are not suitable in machine learning contexts. Another example is dataSHIELD~\citep{Gaye:2014sf} which comprises a technically advanced computational infrastructure and uses distributed computing and parallelized analysis to enable joint analysis, but does not support machine learning models. Federated learning models such as the ones proposed by \cite{Shokri:2015os} for deep learning are not generally applicable to other machine learning methods and also complex to implement.
 
In this manuscript, we propose a light-weight approach to improve predictions over different sources without explicitly sharing the data, by means of aggregating conformal predictions computed at individual locations (see Figure~\ref{fig:overview}b). 
Conformal predictors provide a layer on top of underlying machine learning algorithms, and produce results with valid measures of confidence~\cite{vovk2005algorithmic}.
Here we extend the basic conformal prediction framework to handle multiple data sources and without sharing of data between sources. We propose to aggregate conformal predictions from multiple sources, where transductive conformal predictors (TCP) are applied on the multiple data sources and their individual predictions are aggregated to form a single prediction on a new example.
The advantages of this approach are two-fold: 
Firstly, it extends the existing framework of conformal prediction to multiple data sources that do not require sharing of data between the sources. Secondly, combined learning produces more efficient predictions than individual learning. The method is flexible in the sense that it supports flexible number and sizes of data sources.

%We illustrate the method using simulated and real data sets, and we show that the proposed method produces much more efficient predictions than individual analsys.

Our main research objectives in this paper are summarized as follows:

\begin{enumerate}
\item to investigate if and how the number of data sources and size of the sources affect the aggregated efficiency and validity

\item to evaluate how good both ``aggregated equally partitioned" and ``aggregated randomly partitioned" perform when compared to the whole (pooled) data set.

\item to evaluate if and under what conditions aggregated TCP delivers acceptable results when compared to pooled data

%\item to show that the combined results are better (synergy) than the individual source p-values.

\end{enumerate}

The organization of the paper is as follows. In section 2, we introduce the background concepts and notations used throughout the paper. In Section 3, we introduce the concept of aggregating conformal predictions from multiple sources. In Section 4, we discuss the statistical properties of aggregated conformal predictions from multiple sources. In Section 5, we perform numerical analysis on simulated and real datasets. Finally, in Section 6, a summary of the paper is provided.
%We have also included an appendix that reviews the most relevant aspects about TCP, ICP, CCP and ACP.\todo{do we really have to do this?}

\section{Methods}

\subsection{Conformal prediction}
%This section gives a brief background about TCP and fixes notations and definitions used throughout the paper.

%\subsection{Transductive Conformity Prediction (TCP)}
%\section*{Notations and Assumptions}
The object space is denoted by $\mathcal{X} \in \mathbb{R}^p$, where $p$ is the number of features, and  label space is denoted by $\mathcal{Y} \in \{ 0,1 \}$. We assume that each example consists of an object and its label, and its space is given as $\mathcal{Z} := \mathcal{X} \times \mathcal{Y}$. %are fixed throughout the article. 
The typical classification problem is, given a training dataset $Z = \{ z_1 , ..., z_n \} $ -- where $n$ is the number of examples in the training set, and each example $z_i = (x_i, y_i)$ are labeled examples -- we want to predict the label of a new object $x_{new}$ whose label is unknown. We also assume the exchangeability of examples throughout the paper.
%First, we define a transductive non-conformity measure and transductive conformity score in the following.

The nonconformity measure is the score from a function that measures the strangeness of an example. In this study we use the noncomformity measure from a random forest classifier (RFC) which outputs $\tilde{y}=F_{RF}(x)$ where $\tilde{y} \in [0,1]$. Furthermore the noncomformity scores are for $0$-label $\alpha_{i} = 1-\tilde{y_{i}}$ and the $1$-label $\alpha_{i} = \tilde{y_{i}}$. However, the methodology is general and other underlying machine learning methods and nonconformity scores are equally applicable.To compute the corresponding p-values, we use the smoothed mondrian approach~\citep{vovk2005algorithmic}, where the taxonomy is defined by the labels.

Conformal prediction provides a layer on top of an existing machine learning method and uses available data to determine valid prediction regions for new objects~\citep{vovk2005algorithmic}. 
The predicted region of an object is a subset of $\mathcal{Y}$ , denoted as $\Gamma^{\epsilon} = \{ y \mid p_y > \epsilon \}$, at a significance level $\epsilon$. In the transductive approach (Algorithm \ref{algo:TCP}), the underlying model must be retrained each time an object is to be predicted. For further details on TCP, we refer to \cite{vapnik1998statistical}, \cite{shafer2008tutorial}, \cite{vovk2005algorithmic} and \cite{balasubramanian2014conformal}.

\vspace{0.3cm}
\begin{algorithm}[H]
 \caption{\textbf{Mondrian TCP with random forest}} \label{algo:TCP}
 \textbf{Input:}{ (training dataset:$Z$, object to predict:$x_{new}$, label set:$Y$, a nonconformity measure:$F_{RF}$}\\
 \textbf{Output:}{\textbf{ p-values} }\\
 %\textbf{Initialization\;}
 \For{each $y \in \mathcal{Y}$ }{
 	$z_{n+1} = (x_{new},y) $;\\
 	$Z^* = (Z,z_{n+1})$ ;\\
	Create $F_{RF}$ using $Z^*$\\
	\If{$y_{i}=y$}{
		 \For{each $i \in n+1$ }{
			\textbf{if} $y_{i}=1$ \textbf{then}  $\alpha_i = F_{RF}(x_{i})$\\
			\textbf{if} $y_{i}=0$ :  $\alpha_i = 1- F_{RF}(x_{i})$\\
		 }
		 }
	Compute $p_{y}$ 
%	Compute p-value: $ p(y) = \frac{| \{ i \in \{1,..,n+1\} : y_i=y, \alpha_i(y) < \alpha_{new}(y) \} | + u*| \{ i \in \{1,..,n+1\} : y_i=y, \alpha_i(y) < \alpha_{new}(y) \} |}{n_y}$;\\
  }

% $\textbf{p-values} = \{ p(y)| y \in \mathcal{Y}\}$;\\
 \textbf{return $p_{new}^0,p_{new}^1$};\\
 \end{algorithm}
 \vspace{10pt}
\textbf{ \\}

%The significance is assumed at $\alpha = 0.05$, so we drop the superscript from $\Gamma$ from now on wards. The predictor make an error when it predicts an empty prediction region for a test case $|\Gamma| = 0$.
 
To assess the quality of a conformal predictor we consider validity and efficiency. A predictor makes an error when the predicted region does not contain the true label $ y \not\in \Gamma^{\epsilon}$. Given a training dataset $Z$ and an external test set $T$,  and $|T| = m$. Suppose that the conformal predictor gives prediction regions as $\Gamma_1^{\epsilon}, ...., \Gamma_m^{\epsilon}$, then the error rate is defined as 

\begin{definition}[Error rate]
\begin{align} \label{eq:errorRate}
		ER^{\epsilon} &= \frac{ 1}{m} \sum\limits_{i=1}^{m} \textbf{I}_{ \{y_i \not\in \Gamma_i^{\epsilon} \} },		
\end{align}	
where $y_i$ is the true class label of the $i^{th}$ test case and $\textbf{I}$ is an indicator function. 	
\end{definition}

In the following, we consider a way of assessing validity of a conformal predictor in terms of deviation of observed from expected error. The deviation from exact validity can be computed as the Euclidean norm of the difference of the observed error and the expected error for a given set of predefined significance levels~\citep{carlsson2017comparing}. Let us assume a set of significance levels $\epsilon = \{ \epsilon_1, ..., \epsilon_k \}$, then the formula for the validity can be given as follows.

\begin{definition}[Validity]
\begin{align} \label{eq:validity}
		VAL = \sqrt{ \sum\limits_{i=1}^{k} (ER^{\epsilon_i} -\epsilon_i)^2 }
\end{align}	 
\end{definition}
%We note that TCPs are valid in the sense error rate is less than equal to the significance level, or equivalently the observed error is less than equal to the expected error.

We use observed fuzziness~\citep{vovk2016criteria} as our measure of efficiency,  defined as the sum of all p-values for the incorrect class labels. %For example, for binary classification the efficiency can be computed as follows.
\begin{definition}[Efficiency]
\begin{align} \label{eq:efficiency}
	EFF =\frac{ 1}{m} \sum\limits_{i=1}^{m} \sum\limits_{y_i \neq y }  p_i^y,		
\end{align}
\end{definition}
We note that for the above measure of validity and efficiency, smaller
values are preferable.

\subsection{Non-Disclosed aggregated Conformal Prediction}
%In this section, we present the proposed aggregation method called non-disclosed aggregated conformal prediction. 
%We propose to aggregate conformal p-values across potentially unbalanced data sources, where the number of sources and the size of each data source may vary, and where data is not disclosed between the data sources. 
%
%For illustration purpose, let us consider a binary classification problem, and 
Suppose we have $K$ data sources, each with a training dataset $D_k$ of arbitrary sizes where $k \in {1,...,K}$. For a new object $x_{new}$, the objective is to aggregate p-values at the location $A$ that were computed in each data source using Mondrian TCP with random forest (Algorithm~\ref{algo:TCP}). We name this aggregated predictor Non-Disclosed aggregated Conformal Predictor (NDCP), with details described in Algorithm~\ref{algo:NDCP}. The result is a set of aggregated p-values, where no data training data is disclosed between the data sources and between the data sources and location $A$, but the only information that is transmitted between data sources and $A$ is the object to predict and the resulting p-values.

 \begin{algorithm}[H]
 \textbf{Input:}{ $D_1,...D_K  ;  x_{new}$}\\
 \textbf{Output:}{\textbf{ Aggregated p-values} }\\
 \textbf{Steps\;}
 \For{each $D_k$,  $k \in \{ 1,...,K\} $ }{
	Compute $p^0$ and $p^1$ using Algorithm 1 at each data source\\
	Transfer $p^0$ and $p^1$ to location $A$\\
	$\bar{p}^0 = \bar{p}^0 + p^0$\\
	$\bar{p}^1 = \bar{p}^1 + p^1$\\  
 }
$\bar{p}_{new}^0 = \bar{p}^0 / K$\\   
$\bar{p}_{new}^1 = \bar{p}^1 / K$\\   
 \caption{Non-Disclosed Conformal Prediction (NDCP)} \label{algo:NDCP}
 \textbf{return} $\bar{p}_{new}^0$ , $\bar{p}_{new}^1$
 \end{algorithm}

\section{Evaluation}

We evaluate NDCP on five binary classification datasets from the UCI repository (Table \ref{tab:datasets}) that are randomly partitioned into a training set ($80\%$) and a test set ($20\%$). 

\begin{table} [H]
\caption{Description of the datasets from UCI repository that are used in the evaluation.} \centering
\vspace{1em}
\begin{tabular}{llllc}
\toprule
Dataset && Observations & \# Features \\
\midrule
Spambase (SB) & & 4601 & 57 \\
\hline
Breast Cancer Wisconsin (BC)  & & 699 & 10 \\
\hline
Mushroom (Mush)& & 8124 & 22 \\
\hline
 First-order theorem proving (FOTP) & & 6118 & 51 \\
\hline
Phishing Websites (Phish) & & 2456 & 30 \\
\bottomrule
\end{tabular}\label{tab:datasets}
\end{table}

The steps are described below, and also illustrated in Figure \ref{fig:uACP}.

%Add comments on random forest, what package and version we used

\begin{enumerate}

\item The training data set is randomly split into K parts (disjointly) with varying sizes. For example, Let $Z = \{ z_1 , ..., z_n \} $ be the data set, then we divide the dataset into $S_1, ..., S_K$ such that $Z = \bigcup_{i=1}^K S_i$, $k_i = |S_i|$ and $n = k_1+ ...+k_K$. More specifically, the following types of partitions were made to create different scenarios, with one to compare with unpartitioned data (a TCP trained on all data):

\begin{enumerate}
	\item Pooled source (pooled): Entire training set is considered as one single data source.
	\item Equally sized sources (EqSrc): Training set is randomly partitioned into equally sized sources and  each partition is considered as a proper training set to model and compute p-values, and then p-values are aggregated for all sources. We consider 2, 4 and 6 equal sized sources. %in three different settings.
	\item Unequally sized sources (RandSrc): Training set is randomly partitioned into unequally sized sources and  each partition is considered as a proper training set to model and compute p-values, and then p-values are aggregated for all sources. We consider 2, 4 and 6 unequally sized sources, and we repeat it five times to get five different set of sizes for each source.

\end{enumerate}

\item Predict p-values for the different scenarios using Algorithm \ref{algo:NDCP}

\item Perform evaluation based on validity and efficiency

\item Repeat step 1 to 3  five times.

%\item Pairwise compare the $q$ uACP results obtained to see whether the number of sources or different size of the sources make any difference in validity or efficiency. %(this part is not clear yet).

\end{enumerate}

Results were combined for each scenario over all datasets, and then a pairwise comparison using a Wilcoxon signed-rank test on validity and efficiency for all scenarios were performed.

%The results obtained for $q$ NDCP are then compared pairwise and also compared with the pooled (total) data and equal (balanced) sized partitions. Below we describe the study procedure (random sized or equal sized partition) and the parameters (the number and size of the sources) to be investigated in the result section. 

%In the following section, we show results when the procedure is applied to five different datasets. 
\begin{center}
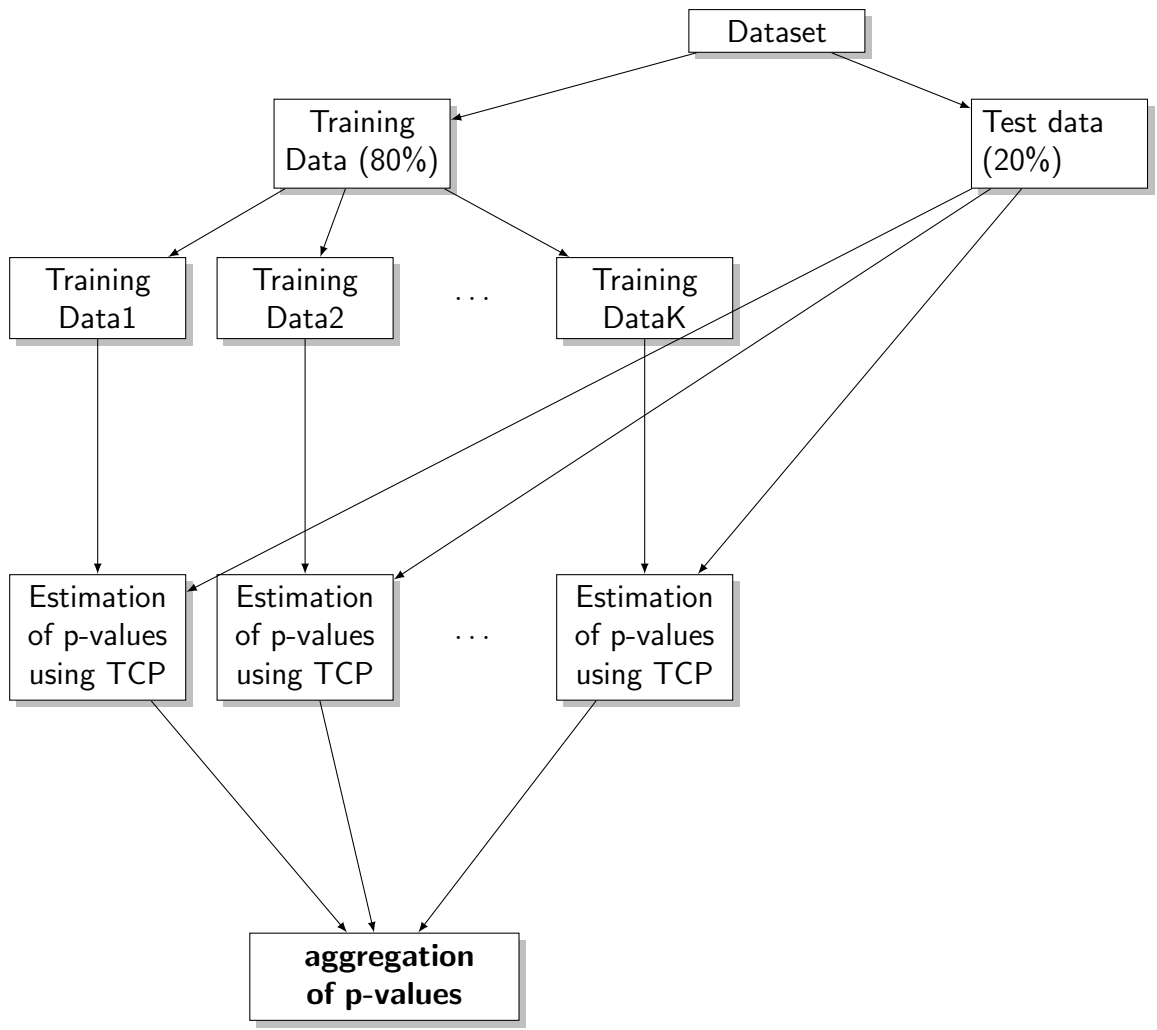
\begin{figure}
\begin{tikzpicture}[
  level 1/.style={sibling distance=40mm},
  edge from parent/.style={->,draw},
  >=latex]

% root of the the initial tree, level 1
\node[root] {Dataset}
% The first level, as children of the initial tree
  child {node[level 2,xshift=-100pt] (c1) {Training Data (80\%)}}
  child {node[level 3, xshift=50pt] (c2) {Test data (20\%)}};

% The second level, relatively positioned nodes
\begin{scope}[every node/.style={level 2}]
\node [below of = c1, xshift=-100pt, yshift=-30] (c11) {Training Data1};
\node [right of = c11, xshift=50pt] (c12) {Training Data2};
\node [right of = c12, xshift=100pt] (c13) {Training DataK};

%\node [below of = c11, yshift=-100] (m11) {Endpoint prediction probabilities using TCP};
%\node [below of = c12, yshift=-100] (m12) {Endpoint prediction probabilities using TCP};
%\node [below of = c13, yshift=-100] (m13) {Endpoint prediction probabilities using TCP};

\node [below of = c11, yshift=-100] (p11) {Estimation of p-values using TCP};
\node [below of = c12, yshift=-100] (p12) {Estimation of p-values using TCP};
\node [below of = c13, yshift=-100] (p13) {Estimation of p-values using TCP};

\node [below of = p12, xshift=30, yshift=-100, text width=8em] (a11) {\textbf{ aggregation of p-values } };

\end{scope}

% lines from each level 1 node to every one of its "children"
%\foreach \value in {1,2,3}
 % \draw[->] (c1.195) |- (c1\value.west);

  \draw[->] (c1)->(c11);
  \draw[->] (c1)->(c12);
  \draw[->] (c1)->(c13);
\path (c12) -- node[auto=false]{\ldots} (c13);

%\draw[->] (c11)->(m11);
%  \draw[->] (c12)->(m12);
%  \draw[->] (c13)->(m13);
 % \path (m12) -- node[auto=false]{\ldots} (m13);

\draw[->] (c11)->(p11);
  \draw[->] (c12)->(p12);
  \draw[->] (c13)->(p13);
  \path (p12) -- node[auto=false]{\ldots} (p13);
  
  \draw[->] (p11)->(a11);
  \draw[->] (p12)->(a11);
  \draw[->] (p13)->(a11);
  
  \draw[->] (c2)->(p11);
  \draw[->] (c2)->(p12);
  \draw[->] (c2)->(p13);
\end{tikzpicture}
\caption{Evaluation of NDCP Algorithm for a given dataset} \label{fig:uACP}
\end{figure}
\end{center}

\newpage
\section{Results and discussion}\label{sec:results}
\begin{figure}[H]
\centering
\begin{subfigure}{\textwidth}
  \centering
  \includegraphics[width=.75\linewidth]{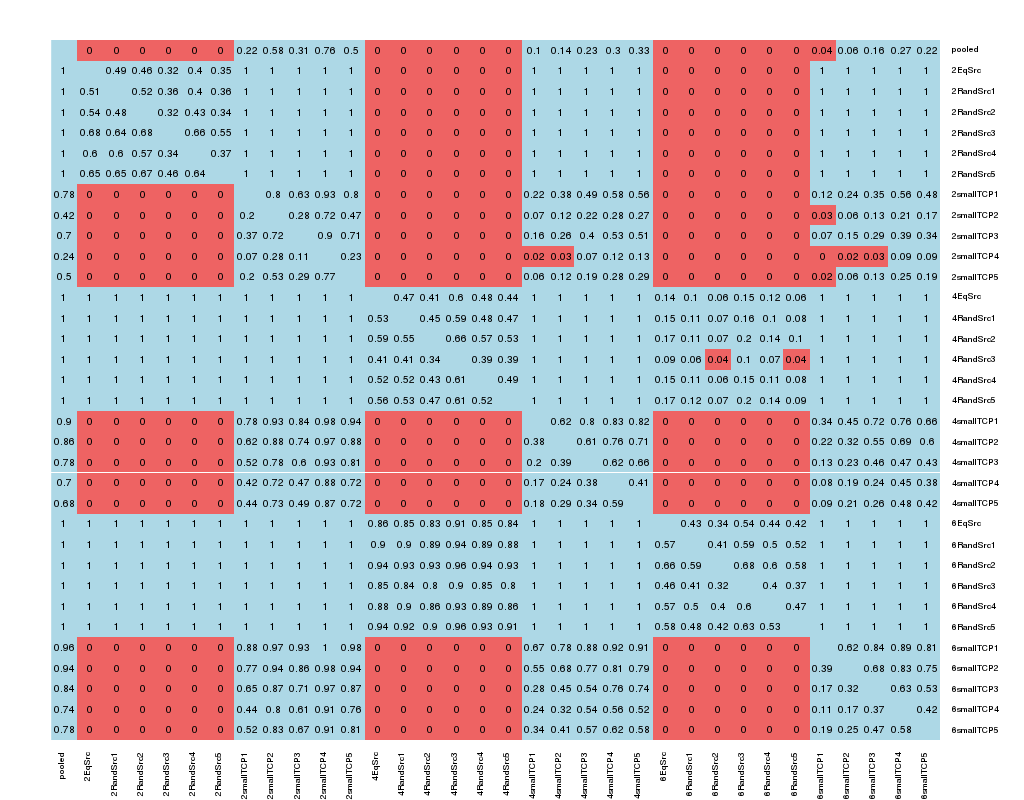}
%  \caption{Validity}  \label{fig:valCombined}
\end{subfigure}%

\begin{subfigure}{\textwidth}
  \centering
  \includegraphics[width=.75\linewidth]{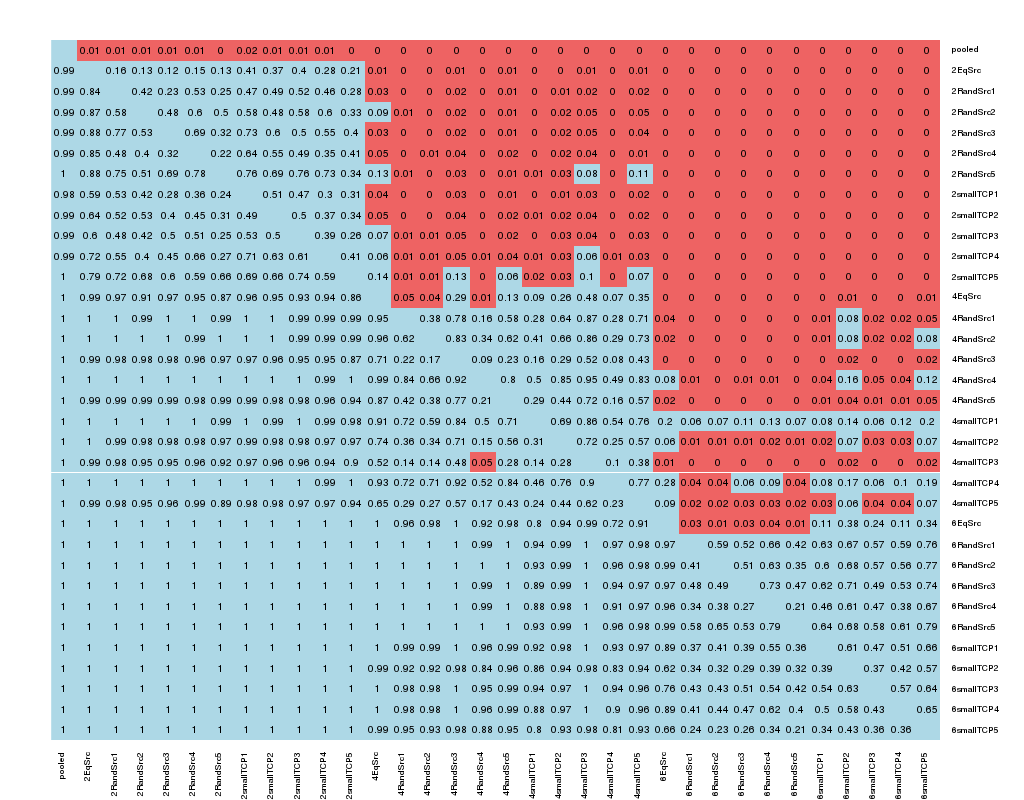}
  %\caption{Efficiency}  \label{fig:effCombined}
\end{subfigure}%
\caption{Results of Wilcoxon signed-rank tests for two alternative hypotheses relating validity (a) and observed fuzziness (b) with combining all the datasets. The p-values are shown for the scenarios in the right column having greater values than the scenarios in the rows. All significant p-values are marked in red. Pooled: Unpartitioned dataset. EqSrc: equally partitioned data sources, RandSrc: randomly partitioned data sources. smallTCP: a single TCP model.} \label{fig:testCombined}
\end{figure}

\begin{figure}[H]
\begin{center}

%\begin{subfigure}{\textwidth}\centering
%  \includegraphics[width=12cm,height=6cm]{images/boxplotCombined}
%  \caption{Validity}  \label{fig:valBC}
%\end{subfigure}%

%\begin{subfigure}{\textwidth} \centering
 
  \includegraphics[scale=0.8]{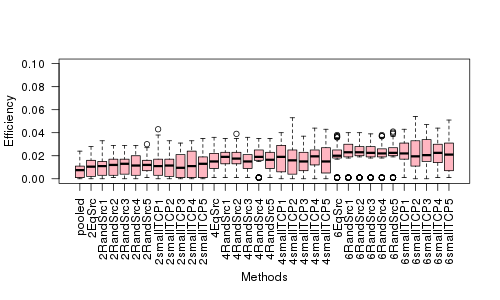}
 % \caption{Efficiency}  \label{fig:effBC}
%\end{subfigure}%
%\caption{Box plot of validity (a) and observed fuzziness (b) with combined data.}
\caption{Box plot of observed fuzziness for aggregating 0, 2, 4, and 6 non-disclosed data sources. Pooled: Unpartitioned dataset. EqSrc: equally partitioned data sources, RandSrc: randomly partitioned data sources. smallTCP: a single TCP model.}
\label{fig:boxplotCombined}
\end{center}
\end{figure}

The results from the pairwise comparison of validity and efficiency (observed fuzziness) are shown in Figure \ref{fig:testCombined}. To illustrate the quantitative difference between the scenarios, box plots are presented in Figure~\ref{fig:boxplotCombined}.
Results in Figure~\ref{fig:testCombined} show that pooled is significantly more efficient than all other models, as would be expected, but in absolute numbers the decrease in efficiency is not so large when using NDCP. When comparing NDCP with individual smallTCP, we do not see a significant improvement in efficiency using NDCP but we observe a reduced variance, consistent with previous work~\citep{Carlsson:2014qr}, when there are 4 or more partitions. We also observe in Figure~\ref{fig:testCombined} that there is no significant difference between 'aggregated equally partitioned' and 'aggregated randomly partitioned', which would make the method generally applicable regardless of the sizes of individual training sets.

%NDCP more efficient predictions than individual analyses
%1. To investigate if and how the number of data sources and size of the sources affect the aggregated efficiency and validity

\begin{figure}[H]
\begin{center}
  \begin{subfigure}{.3\textwidth}
  \centering
  \includegraphics[scale=0.2]{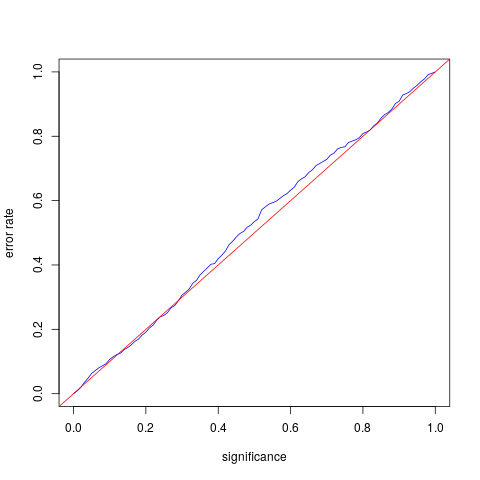}
  \caption{pooled} \label{fig:pooledCalibrationPlot}
   \end{subfigure}    
  \begin{subfigure}{.3\textwidth}
  \centering
  \includegraphics[scale=0.2]{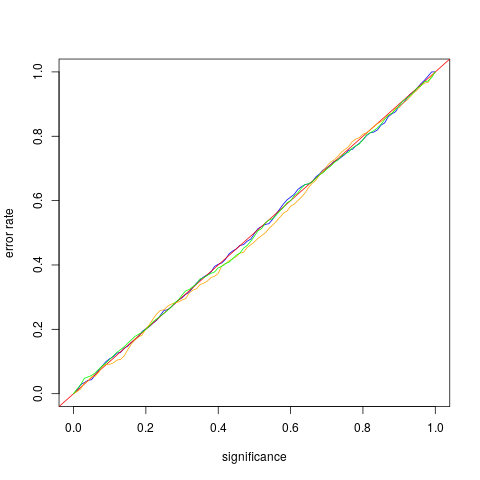}
  \caption{smallTCPs}\label{fig:valIndividual}
  \end{subfigure}
  \begin{subfigure}{.3\textwidth}
  \centering
  \includegraphics[scale=0.2]{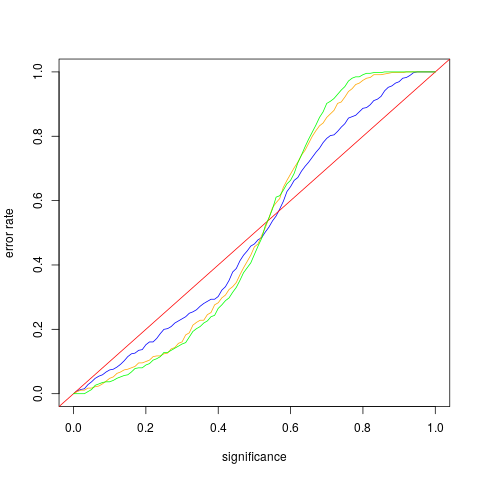}
  \caption{NDCPs}\label{fig:valCombined}
  \end{subfigure}
  
 \caption{Calibration plot for various models. a) Calibration plot of TCP for one fold of Spambase dataset. b) Calibration plot of randomly partitioned small TCPs for one fold of  Spambase dataset. Blue, orange and  green line indicate each small TCP from two, four and six source random partitions respectively. c) Calibration plot of NDCPs for one fold of Spambase dataset. Blue, orange and  green line indicate NDCP from two, four and six source random partitions respectively}
 
\end{center}
\end{figure}

%\subsection{Validity}
Regarding validity, we observe that the pooled model is always valid, as an example see Figure~\ref{fig:pooledCalibrationPlot} for the Spambase dataset. Further, we see that individual small models are also valid, see Figure~\ref{fig:valIndividual} for randomly partitioned small TCPs for Spambase dataset. Consistent with previous work by Linusson et al~\cite{Linusson:2017dn} and Carlsson et al.~\cite{Carlsson:2014qr}, NDCP is less valid overall, see Figure~\ref{fig:valCombined} for randomly partitioned NDCPs for Spambase dataset, but calibration plot shows conservative validity for the significance levels 0 to 0.5 which is the interesting region for predictions. This is a known issue that requires further research; we settle here with the observation that validity does not seem to be a practical problem for NDCP in the interesting significance region.

%Single source, 2EqualSizedSource, 4EqualSizedSource, 6EqualSizedSource, 2UnequalSizedSource, 4UnequalSizedSource and 6UnequalSizedSource. The details of the datasets used and the results on individual datasets are given in Supplementary Material.%are given in the following, these data have been taken from the UCI Repository of machine learning batabases: ftp://ftp.ics.uci.edu/pub/machine-learning-databases/.
%
%
%\subsection{Empirical results on Combined data}

%Another figure

%\section{Results and Discussions}
%The aim of this study was to improve predictions over different data sources without explicitly sharing the data, by aggregating conformal predictions computed at individual locations. In order to do so, we investigated if and how the number of data sources and size of these affect the aggregated efficiency and validity.

%\subsection{Efficiency}
%2. to evaluate how good both aggregated equally partitioned and aggregated randomly partitioned perform when compared to the whole (pooled) data set.

%3. to evaluate if and under what conditions unbalanced aggregated TCP delivers acceptable results when compared to pooled data

\section{Conclusions}
We present a method to aggregate conformal predictions from multiple sources while preserving data privacy. The method is a generalization of the basic conformal prediction framework to handle multiple data sources without disclosing data between the data sources. Due to its low complexity for implementation, we believe the method will be useful for organizations that wish to make predictions over combined data without disclosing data to each other, such as for drug discovery problems when pharmaceutical companies wishes to establish predictive models of drug safety.

\bibliographystyle{elsarticle-harv} 
\bibliography{NDACP}
\end{document}